\documentclass[letterpaper]{article} 
\usepackage{aaai25}  
\usepackage{times}  
\usepackage{helvet}  
\usepackage{courier}  
\usepackage[hyphens]{url}  
\usepackage{graphicx} 
\usepackage{natbib}  
\usepackage{caption} 
\usepackage{algorithm}
\usepackage{algorithmic}
\usepackage{bm}
\usepackage{amsmath}
\usepackage{amssymb}
\usepackage{xcolor} 
\usepackage{booktabs}       
\usepackage{multirow}

\usepackage{newfloat}
\usepackage{listings}
\DeclareCaptionStyle{ruled}{labelfont=normalfont,labelsep=colon,strut=off} 
\floatstyle{ruled}
\newfloat{listing}{tb}{lst}{}
\floatname{listing}{Listing}
\title{Debiased Multimodal Understanding for Human Language Sequences}
\author{
    Zhi Xu\textsuperscript{\rm 1}\equalcontrib, Dingkang Yang\textsuperscript{\rm 1}\equalcontrib\thanks{Project lead.}, Mingcheng Li\textsuperscript{\rm 1}, Yuzheng Wang\textsuperscript{\rm 1}, Zhaoyu Chen\textsuperscript{\rm 1}, \\ Jiawei Chen\textsuperscript{\rm 1},   
    Jinjie Wei\textsuperscript{\rm 1}, Lihua Zhang\textsuperscript{\rm 1,2,3}\thanks{Corresponding author.} \\
}
\affiliations{
    \textsuperscript{\rm 1}Academy for Engineering and Technology, Fudan University\\
    \textsuperscript{\rm 2}Institute of Metaverse \& Intelligent Medicine, Fudan University, Shanghai, China\\
    \textsuperscript{\rm 3}Cognition and Intelligent Technology Laboratory, Shanghai, China\\
     \{zhixu20, dkyang20\}@fudan.edu.cn
}

\usepackage{bibentry}

\begin{document}

\maketitle

\begin{abstract}
Human multimodal language understanding (MLU) is an indispensable component of expression analysis (\textit{e.g.}, sentiment or humor) from heterogeneous modalities, including visual postures, linguistic contents, and acoustic behaviours. Existing works invariably focus on designing sophisticated structures or fusion strategies to achieve impressive improvements. Unfortunately, they all suffer from the subject variation problem due to data distribution discrepancies among subjects. Concretely, MLU models are easily misled by distinct subjects with different expression customs and characteristics in the training data to learn subject-specific spurious correlations, limiting performance and generalizability across new subjects.
Motivated by this observation, we introduce a recapitulative causal graph to formulate the MLU procedure and analyze the confounding effect of subjects. Then, we propose SuCI, a simple yet effective causal intervention module to disentangle the impact of subjects acting as unobserved confounders and achieve model training via true causal effects. As a plug-and-play component, SuCI can be widely applied to most methods that seek unbiased predictions. Comprehensive experiments on several MLU benchmarks clearly show the effectiveness of the proposed module.
\end{abstract}

\section{Introduction}
\label{sec:intro}
As a research hotspot that combines linguistic and non-verbal behaviours (\textit{e.g.}, visual and acoustic modalities), human multimodal language understanding (MLU) has attracted significant attention from computer vision \cite{yang2022disentangled}, natural language processing \cite{tian2022debiasing}, and speech recognition communities \cite{yang2022contextual, yang2022learning} in recent years. Thanks to the progressive development of multimodal language benchmarks \cite{hasan2019ur,zadeh2018multimodal, zadeh2016mosi}, extensive studies \cite{rahman2020integrating,han2021improving,liu2018efficient,yu2021learning, tsai2019multimodal,liang2021attention,lv2021progressive,pham2019found,sun2022cubemlp,lei2023text}  have 
presented impressive multimodal models on training data containing distinct subjects, diverse topics, and different modalities. 
Despite the achievements of previous approaches by exploiting representation learning architectures  \cite{yang2022disentangled,liang2021attention} and fusion strategies \cite{tsai2019multimodal,lv2021progressive}, they invariably suffer from a prediction bias when applied to testing samples of new subjects.

\begin{figure*}[t]
\centering
\includegraphics[width=0.9\linewidth]{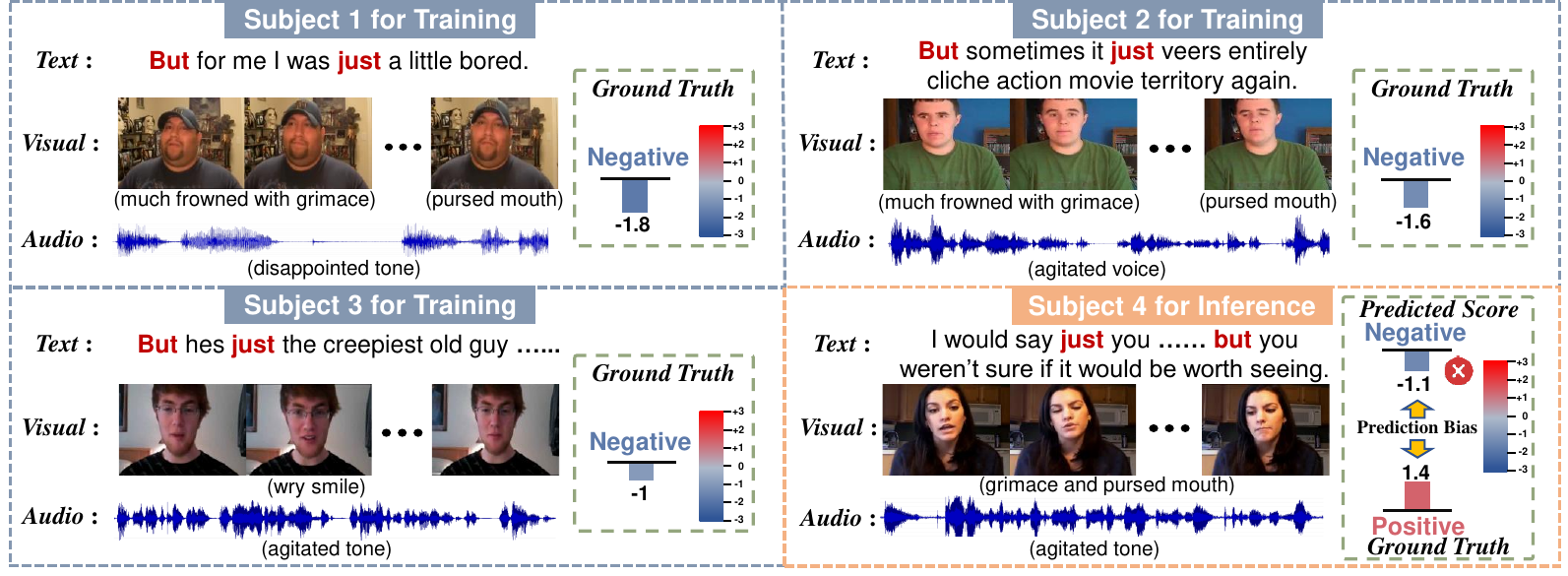}
  \caption{Examples on the MOSI benchmark illustrate the subject variation problem. Multimodal expressions from four subjects potentially convey distinct semantic correlations due to their different customs and styles in expressing sentiments. 
  }
  \label{intro}
\end{figure*}

The harmful prediction bias is mainly caused by the subject variation problem. Specifically, different subjects' expression styles and behaviours (\textit{e.g.}, facial expressions or acoustic information) from the training data are highly idiosyncratic in social communication, affected by the subjects' customs or culture~\cite{li2020deep}.
Once well-designed models are trained on such data, subject-specific semantic correlations (\textit{e.g.}, particular facial action unit co-occurrence \cite{chen2022causal}) would inexorably affect performance and generalizability.
Worse still, the spurious connections between the trained models and specific subjects will be transmitted via multiple modalities when the data paradigm is extended from isolated to multimodal situations.
Recall the prominent MLU benchmarks (\textit{e.g.}, MOSI \cite{zadeh2016mosi} for sentiment analysis or UR\_FUNNY~\cite{hasan2019ur} for humor detection), whose collectors advocated video-level data splitting so that segments from the same video will not appear across train, valid, and test splits. Although the trained models may avoid memorizing the average affective state of a subject \cite{liang2021multibench}, they cannot generalize well across new subjects. The examples in Figure~\ref{intro} provide strong evidence of this.
Concretely, subjects 1, 2, and 3 tend to use sentimentally unimportant words ``\emph{but}'' and ``\emph{just}'' to express negative emotions.
In this case, the MLU model \cite{li2023decoupled} is misled to focus on spurious clues from the textual utterances of the subjects and make an entirely incorrect prediction when applied to subject 4. Similar observations are found in the visual and audio modalities. For instance, the trained model erroneously takes subject-specific facial appearances (\textit{e.g.}, ``grimace and pursed mouth'' from subjects 1 and 2) and acoustic behaviours (\textit{e.g.}, ``agitated tone'' from subjects 2 and 3) as semantic shortcuts to infer frustrating negative sentiment.

Motivated by the above observations, this paper aims to improve MLU methods by causal demystification rather than beating them. We propose a Subject Causal Intervention module (SuCI) to disentangle the impact of semantic differences among subjects from multimodal expressions. Specifically, we first formulate a universal causal graph to analyze the MLU procedure and interpret the causalities among the variables. From the causal inference perspective, the \textit{subjects} are essentially regarded as \textit{confounders}~\cite{glymour2016causal}, which mislead the models to learn subject-specific semantic correlations in the training data, as well as causing prediction bias for new subjects in the inference phase.
For clarity, we represent the multimodal inputs as $\bm{X}$ (cause) and its corresponding predictions as $\bm{Y}$ (effect). Existing MLU models aim to approximate $P(\bm{Y}|\bm{X})$ as much as possible while failing to perform unbiased predictions. Unlike conventional likelihood estimation $P(\bm{Y}|\bm{X})$, our SuCI achieves subject de-confounded training and removes subject-caused confounding effects by embracing causal intervention $P(\bm{Y}|do(\bm{X}))$ regarding the backdoor adjustment rule~\cite{pearl2009causal}. As a model-agnostic component, SuCI can be readily integrated into most MLU models to perform unbiased predictions by pursuing true causal effects among variables. 

To sum up, this paper has the following contributions. \textbf{(i)} We investigate the subject variation problem in MLU tasks via a tailored causal graph and identify the subjects as confounders which misleads the models to capture subject-specific spurious correlations and cause prediction bias. \textbf{(ii)} Based on the causal theory of backdoor adjustment, we present SuCI, a subject causal intervention module to remove prediction bias and confounding effects of subjects. \textbf{(iii)}
We evaluate the
effectiveness of SuCI on several MLU benchmarks. Experimental results
show that SuCI can significantly and consistently
improve existing baselines, achieving new SOTA performance.

\section{Related Work}
\textbf{Human Multimodal Language Understanding.}
Benefiting from available human communication resources and data, MLU benchmarks~\cite{zadeh2018multimodal,zadeh2016mosi,hasan2019ur} with different scales and typologies have been increasingly developed and applied in recent years.
Recent MLU tasks focus on subject-centered intention understanding and behavior analysis from text, visual, and audio modalities, including but not limited to emotion recognition \cite{lv2021progressive}, sentiment analysis \cite{han2021improving}, and humor detection \cite{hasan2019ur}.
Considering the heterogeneous nature of multimodal languages, numerous works have presented seminal network structures \cite{pham2019found,liang2021attention, yu2021learning,sun2022cubemlp}, fusion strategies \cite{tsai2019multimodal,tsai2018learning,zadeh2017tensor,rahman2020integrating}, and representation learning paradigms \cite{yang2022disentangled,yang2022learning,yang2022emotion,li2023towards}. For instance, 
MFSA~\cite{yang2022learning} presented a factorized representation strategy to learn similarities and differences among multimodal languages.
Despite achievements, existing methods invariably suffer from performance bottlenecks due to subject-related prediction bias.
In comparison,
we identify subjects in MLU benchmarks as harmful confounders from a causal perspective and improve different models with our SuCI.

\textbf{Causal Demystification.}
Causal demystification as a potential statistical theory aims to pursue causal effects among observed variables rather than their shallow correlations. Benefiting from the advances in learning-based technologies~\cite{yang2023what2comm,yang2023aide,yang2023spatio,yang2023how2comm,liu2023stochastic,liu2023generalized}, several studies exploring causality are mainly divided into two channels: causal intervention~\cite{lin2022causal,yang2021causal} and counterfactual reasoning~\cite{qian2021counterfactual,tang2020unbiased}. Intervention~\cite{pearl2009causal} focuses on altering the natural tendency of the independent variable to vary with other variables to eliminate the impact of adverse effects. Counterfactuals depict imagined outcomes produced by factual variables under different treatment conditions~\cite{pearl2009causality}. Given the unbiased estimations provided by causal inference, it is increasingly applied to various downstream tasks such as computer vision~\cite{yang2021deconfounded,yang2023context}, natural language processing~\cite{zhang2021biasing,huang2020counterfactually}, and reinforcement learning~\cite{dasgupta2019causal}.
In this paper, we address the confounding effect from multiple modalities by embracing causal intervention, which is more generalizable and adaptable for multimodal approaches.

\section{Methodology}

\subsection{Structural Causal Graph in MLU Tasks}

\begin{figure}[t]
  \centering
  \includegraphics[width=0.7\linewidth]{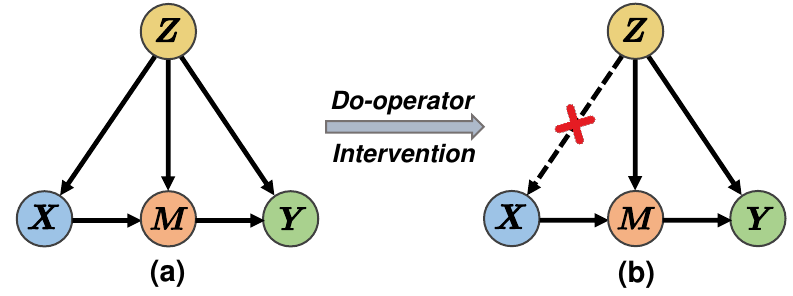}
  \caption{The causal graph explains causal effects of MLU procedure. Nodes denote variables and arrows denote the direct causal effects. (a) The
conventional likelihood estimation $P(\bm{Y}|\bm{X})$. (b) The causal intervention $P(\bm{Y}|do(\bm{X}))$.
  }
\label{graph}
\end{figure}

To systematically diagnose the confounding effect present in MLU tasks, we first design a recapitulative causal graph to summarize the causal relationships among variables. The mainstream graphical notation following the structured causal model~\cite{pearl2000models} is adopted due to its intuitiveness and interpretability. Concretely, a causal graph $\mathcal{G} = \{ \mathcal{N}, \mathcal{E} \}$ is considered a directed acyclic graph, which represents how a set of variables $\mathcal{N}$ convey causal effects through the causal links $\mathcal{E}$. As shown in Figure~\ref{graph}(a), there are four variables in the MLU causal graph, including multimodal inputs $\bm{X}$, multimodal representations $\bm{M}$, unintended confounders $\bm{Z}$, and predictions $\bm{Y}$. Note that our causal graph applies to various MLU approaches since it is highly general, imposing no constraints on the detailed implementations.
The causalities are shown below.

\noindent $\blacktriangleright$ \textbf{Link} $ \bm{Z} \rightarrow \bm{X}$. Multimodal expressions from distinct subjects are recorded to produce multimodal inputs $\bm{X}$, where $\bm{X}$ is a generalized definition of text $\bm{X}_{t}$, visual $\bm{X}_{v}$, and audio $\bm{X}_{a}$ modalities for simplicity, \textit{i.e.}, $\bm{X} = \{\bm{X}_{t}, \bm{X}_{v}, \bm{X}_{a}\}$.
Subjects are identified as harmful confounders $\bm{Z}$ due to the subject-related prediction bias caused by different human expression customs and differences~\cite{hasan2019ur,zadeh2018multimodal}. In this case, $\bm{Z}$ is a collective denomination of multimodal confounding sources.
For training inputs $\bm{X}$, $\bm{Z}$ determines the subject-related biased content that is recorded,  \textit{i.e.}, $ \bm{Z} \rightarrow \bm{X}$.

\noindent $\blacktriangleright$ \textbf{Link} $ \bm{Z} \rightarrow \bm{M} \leftarrow \bm{X}$. $\bm{M}$ denotes the refined multimodal representations extracted by any MLU model, which acts as a mediator before the final classifier.
The link $\bm{Z} \rightarrow \bm{M}$ indicates detrimental $\bm{Z}$ confounding models to capture subject-specific characteristics embedded in $\bm{M}$ to produce spurious semantic correlations.
Several intuitive examples are the semantically unimportant words from the text modality and the agitated tones from the audio modality in Figure~\ref{intro}.
Furthermore, $\bm{M}$ contains universal multimodal feature semantics from $\bm{X}$ that can be reflected via the causal link $ \bm{X} \rightarrow \bm{M}$.

\noindent $\blacktriangleright$ \textbf{Link} $ \bm{M} \rightarrow \bm{Y} \leftarrow \bm{Z}$.
The causal path $ \bm{M} \rightarrow \bm{Y} $ reveals that the impure $\bm{M}$ confounded by $\bm{Z}$ further impacts the final predictions $\bm{Y}$ of downstream MLU tasks. Meanwhile, adverse confounders' prior information in the training data implicitly interferes with $\bm{Y}$ along the link $ \bm{Z} \rightarrow \bm{Y} $.

According to the causal theory~\cite{pearl2009causal}, the confounders $\bm{Z}$ are the common cause of $\bm{X}$ and corresponding predictions $\bm{Y}$. The positive effect of the subject-agnostic multimodal semantics provided by $\bm{M}$ follows the desired causal path $ \bm{X} \rightarrow \bm{M} \rightarrow \bm{Y} $, which we aim to achieve and pursue. Unfortunately, $\bm{Z}$ causes subject-related prediction bias and misleads trained models to learn subject-specific misleading semantics rather than pure causal effects, leading to biased predictions on uninitiated subjects. The detrimental effects follow the backdoor paths $ \bm{X} \leftarrow \bm{Z} \rightarrow \bm{Y}$ and $ \bm{X} \leftarrow \bm{Z} \rightarrow \bm{M} \rightarrow \bm{Y}$.

\subsection{Causal Intervention via Backdoor Adjustment}
\label{sec3.2}
Following the causal graph in Figure~\ref{graph}(a), the MLU model relies on the likelihood $P(\bm{Y}|\bm{X})$ for predictions that suffer from backdoor effects, which can be decomposed by the Bayes rule as follows:
\begin{equation}
\small
P(\bm{Y}|\bm{X})=\sum_{\bm{z}}^{} P(\bm{Y}|\bm{X}, \bm{M} = \mathcal{F}_{m}(\bm{X}, \bm{z})) P(\bm{z}|\bm{X}),
\label{one}
\end{equation}
where $\mathcal{F}_{m}(\cdot)$ denotes any vanilla MLU model to learn the multimodal representations $\bm{M}$. $\bm{z}$ is a stratum of confounders (\textit{i.e.}, a subject), which introduces the observational bias via $P(\bm{z}|\bm{X})$.
Theoretically, an ideal solution would be to collect massive data samples to ensure that subjects with different expression characteristics are included in the training and testing sets.
However, this way is unrealistic due to social ethics issues~\cite{jones1999bounded}. 
To address this, we embrace causal intervention $P(\bm{Y}|do(\bm{X}))$ to interrupt the adverse effects propagating between $\bm{X}$ and $\bm{Y}$ along the backdoor paths via the backdoor adjustment theory~\cite{pearl2009causal}.
The $do(\cdot)$ operator is an efficient approximation to implement the empirical intervention~\cite{glymour2016causal}. In our case, backdoor adjustment means measuring the causal effect of each stratum in the subject confounders and then performing a weighted integration based on the prior proportions of samples from different subjects in the training data to estimate the average causal effect. From Figure~\ref{graph}(b), the impact from $\bm{Z}$ to $\bm{X}$ is cut off since the model would enable the subject prototype as the confounder in each stratum to contribute equally to the predictions $\bm{Y}$ by $P(\bm{Y}|do(\bm{X}))$. Eq.~(\ref{one}) with the intervention is formulated as:
\begin{equation}
\small
P(\bm{Y}|do(\bm{X}))=\sum_{\bm{z}}^{} P(\bm{Y}|\bm{X}, \bm{M} = \mathcal{F}_{m}(\bm{X}, \bm{z})) P(\bm{z}).
\label{two} 
\end{equation}
The model is no longer disrupted by subject-specific spurious correlations in backdoor paths since $\bm{z}$ no longer affects $\bm{X}$.
$P(\bm{z})$ is the prior probability that depicts the proportion of each $\bm{z}$ in the whole.

\begin{figure*}[t]
  \centering
      \includegraphics[width=0.8\linewidth]{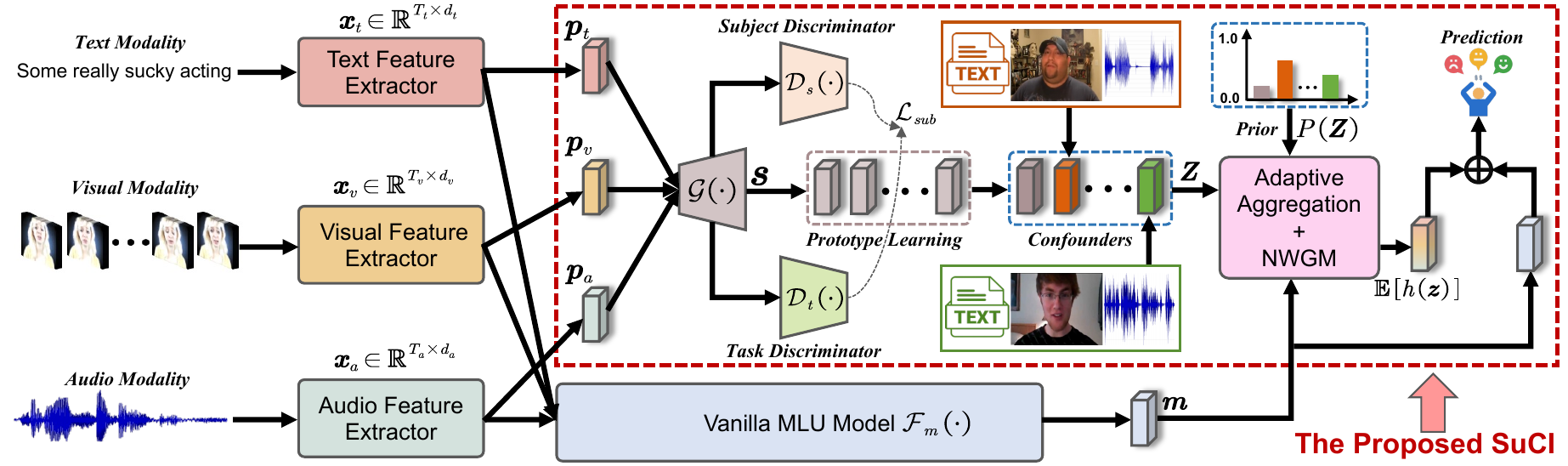}
  \caption{A general MLU pipeline for the subject de-confounded training. The red dotted box shows the core component that achieves the approximation to causal intervention: our SuCI. SuCI can be readily integrated into the vanilla MLU model via backdoor adjustment to mitigate subject-specific spurious correlations and achieve debiased predictions in downstream tasks.
  }
  \label{arc}  
\end{figure*}

\subsection{Subject De-confounded Training with SuCI}
We present a plug-in Subject Causal Intervention module (SuCI) to convert the theoretical intervention in Eq.~(\ref{two}) into a practical implementation.
As Figure~\ref{arc} shows, SuCI can be readily integrated into the vanilla MLU model to estimate $P(\bm{Y}|do(\bm{X}))$ through the subject de-confounded training. The implementation details are as follows.

\noindent \textbf{Subject-specific Feature Disentanglement.}
How to effectively disentangle subject-specific features from heterogeneous modalities is key to determining confounders. Considering that subject-related spurious semantics are usually distributed among different frames of multimodal sequences~\cite{yang2022disentangled}, we first devise a dynamic fusion mechanism to aggregate the frame-level semantic clues. Let the text, visual, and audio modality sequences from the corresponding feature extractors be denoted as $\bm{x}_t\in \mathbb{R}^{T_t \times d_t}$, $\bm{x}_v\in \mathbb{R}^{T_v \times d_v}$, and $\bm{x}_a \in \mathbb{R}^{T_a \times d_a}$, respectively, where $T_{(\cdot)}$ is the frame length and  $d_{(\cdot)}$ is the embedding dimension. The dynamic fusion procedure is formulated as follows:
\begin{align}
\small
\xi_{m} &= \phi (\bm{x}_{m}\bm{w}_{x_m} + \bm{b}_{x_m}) \in \mathbb{R}^{T_m \times 1}, \\
\bm{p}_{m} &= \xi_{m}^{T}\bm{x}_{m} \in \mathbb{R}^{d_m},
\end{align}
where  $ m \in \{ t, v, a\}$, $\phi(\cdot)$ is the softmax function, $\bm{w}_{x_m} \in \mathbb{R}^{d_m \times 1}$, and $\bm{b}_{x_m} \in \mathbb{R}^{T_m \times 1}$ are the learnable parameters. The attention vectors $\xi_{m}$ adaptively capture salient semantics and produce informative representations $\bm{p}_{m}$ based on the contributions of different frames.
Subsequently, we introduce an adversarial strategy to disentangle the subject-specific semantics and avoid the task-related semantics. The design philosophy is to distill pure subject features for better confounder construction. Specifically, a subject generator $\mathcal{G}(\cdot;\theta_{\mathcal{G}})$ is presented to project $\bm{p}_{t}$, $\bm{p}_{v}$, and $\bm{p}_{a}$ of multimodal utterances from the subject into a common space to yield a subject-specific feature $\bm{s}$, expressed as follows:
\begin{equation}
\small
\bm{s} = \mathcal{G}([\bm{p}_{t}, \bm{p}_{v}, \bm{p}_{a}];\theta_{\mathcal{G}}) \in \mathbb{R}^{d_s \times 1},
\end{equation}
where $[\,,]$ stands for the concatenation operator and $d_s = d_t + d_v + d_a$, The generator is implemented as a two-layer perceptron with GeLU~\cite{hendrycks2016gaussian} activation.
$\bm{s}$ is distilled by the following objective:
\begin{equation}
\small
\mathcal{L}_{sub} = \mathcal{CE}(\mathcal{D}_{s}(\bm{s};\theta_{\mathcal{D}_{s}}), y_s) + \mathcal{MSE}(\mathcal{D}_{t}(\bm{s};\theta_{\mathcal{D}_{t}}), \frac{1}{C_t}),
\label{six}
\end{equation}
where $\mathcal{CE}(\cdot)$ and $\mathcal{MSE}(\cdot)$ represent the cross-entropy loss and mean squared error loss, respectively. $\mathcal{D}_{s}(\cdot;\theta_{\mathcal{D}_{s}})$ and $\mathcal{D}_{t}(\cdot;\theta_{\mathcal{D}_{t}})$ are the subject discriminator and task discriminator parameterized by $\theta_{\mathcal{D}_{s}}$ and $\theta_{\mathcal{D}_{t}}$, respectively, which consist of feed-forward neural layers. $y_s$ is the subject label determined by the dataset's subject index. $C_t$ is the number of categories in downstream tasks.
In Eq.~(\ref{six}), $\bm{s}$ is supervised to encourage the output logits of the task discriminator to be equally distributed among all prediction categories to exclude task-related semantic information.  Also, the subject discriminator is optimized to ensure that $\bm{s}$ belongs to a given subject, \textit{i.e.}, contains subject-specific semantics.

\noindent \textbf{Confounder Construction.}
Confounder construction aims to make the model measure the causal effect of subject confounders among different strata during training to avoid subject-related prediction bias.
Considering that subject features are similar within the same stratum and different across strata~\cite{pearl2009causal}, we utilize all the features from a specific subject as the subject prototype, which represents the universal attributes of confounders in a specific stratum.
Concretely, we maintain a memory cache for each subject during training to store and update the subject prototype as a confounder, which is computed as $ {\bm{z}_{i} = \frac{1}{N_i}  \textstyle \sum_{k=1}^{N_i}} \bm{s}_{k}^i$. $N_i$ is the number of training samples for the $i$-th subject, and  $\bm{s}_{k}^i$ denotes the $k$-th feature of the $i$-th subject.
$\bm{z}_{i}$ is updated at the end of each epoch. In practice, each $\bm{z}_{i}$ is initialized by the uniform distribution to ensure stable training at the first epoch. Based on the number of subjects $N_c$, a stratified confounder dictionary is constructed, which is formulated as $\bm{Z}= [\bm{z}_1,\bm{z}_2, \dots,\bm{z}_{N_c} ]$.

\noindent \textbf{Intervention Instantiation.}
Estimating $P(\bm{Y}|do(\bm{X}))$ in practice is high overhead since it requires the forward computation of $\bm{X}$ and $\bm{z}$ for each pair.
To reduce the overhead, we introduce the Normalized Weighted Geometric Mean (NWGM)~\cite{xu2015show} to achieve feature-level approximation for intervention instantiation:
\begin{equation}
\small
P(\bm{Y}|do(\bm{X})) \overset{{\text{NWGM}}}{\approx} P(\bm{Y} | \bm{X}, \bm{M}= \sum_{\bm{z}}^{} \mathcal{F}_{m}(\bm{X}, \bm{z}) p(\bm{z})).
\label{seven}
\end{equation}
In this case, causal intervention makes $\bm{X}$ contribute fairly to the predictions of $\bm{Y}$ by incorporating every $\bm{z}$.
We parameterize a network model to approximate Eq.~(\ref{seven}) as follows:
\begin{equation}
\small
P(\bm{Y}|do(\bm{X})) = \bm{W}_{m} \bm{m} + \bm{W}_h \mathbb{E}[h(\bm{z})],
\end{equation}
where $\bm{W}_{m} \in \mathbb{R}^{d_{h} \times d } $ and $\bm{W}_{h} \in \mathbb{R}^{d_{h} \times d_s } $ are the learnable parameters.             
$\bm{m} \in \mathbb{R}^{d \times 1 }$ is the multimodal representation produced by the vanilla MLU model.
The above approximation is reasonable since the effect on $\bm{Y}$ is attributed to $\bm{M}$ and $\bm{Z}$ in the MLU causal graph. Then, $\mathbb{E}[h(\bm{z})]$ is approximated as an adaptive aggregation for all confounders according to backdoor adjustment, which is formulated as:
\begin{equation}
\small
\mathbb{E}[h(\bm{z})] = \sum_{i=1}^{N_c} \psi_{i} \bm{z}_i p(\bm{z}_i), 
\end{equation}
\begin{equation}
\small
 \psi_{i} = \phi(\frac{(\bm{W}_{q} \bm{m})^{T} (\bm{W}_{k} \bm{z}_{i})}
    {\sqrt{d_s}}),
\end{equation}
where $\bm{W}_{q} \in \mathbb{R}^{d_{n} \times d} $ and $\bm{W}_{k} \in \mathbb{R}^{d_{n} \times d_s} $ are mapping matrices. $p(\bm{z}_i) = \frac{N_i}{N}$, where $N$ is the number of training samples. 
In practice, $\bm{m}$ from one sample queries each $\bm{z}_{i}$ in the confounder dictionary $\bm{Z} \in \mathbb{R}^{N_c \times d_s}$ to obtain the sample-specific attention set $\{\psi_{i}  \}_{i=1}^{N_c}$. The intuition insight is that samples from one subject are impacted by varying degrees of confounder $\bm{z}_{i}$ of other subjects.

\noindent \textbf{Objective Optimization.}
The standard cross-entropy loss is employed as the task-related training objective, which is expressed as $ \mathcal{L}_{task} = - \frac{1}{N_{i}} \sum_{i=1}^{N_{i}} y_t \cdot log \hat{y}_t $, where $y_t$ is the task-related ground truth. The overall objective function is computed as $\mathcal{L}_{all} = \mathcal{L}_{sub} + \mathcal{L}_{task}$.

\section{Experiments}

\begin{table}[t]
\setlength{\tabcolsep}{12pt}
\renewcommand{\arraystretch}{0.8}
\centering
\resizebox{0.8\linewidth}{!}{%
\begin{tabular}{c|ccc}
\toprule
\textbf{Methods}        & $Acc_7 \uparrow$ (\%) & $Acc_2 \uparrow$ (\%) & $ F1 \uparrow$ (\%)   \\ \midrule
EF-LSTM        & 33.7 & 75.3 & 75.2 \\
LF-LSTM        & 35.3 & 76.8 & 76.7 \\ 
Graph-MFN      & 29.6 & 75.4 & 76.6 \\
TFN           & 32.1 & 73.9 & 73.4 \\
LMF          & 32.8 & 76.4 & 75.7 \\
MFM            & 36.2 & 78.1 & 78.1 \\
RAVEN          & 33.2 & 78.0   & 76.6 \\
MCTN          & 35.6 & 79.3 & 79.1 \\
TCSP         & -   & 80.9 & 81.0   \\
PMR          & 40.6 & 83.6 & 83.4 \\
FDMER        & 42.1 & 84.2 & 83.9 \\ \midrule
MulT          & 39.5 & 82.6 & 82.5 \\
MulT + \textbf{SuCI}    & \textbf{40.7}$^{\textcolor{red}{+1.2}}$ & \textbf{83.4}$^{\textcolor{red}{+0.8}}$ & \textbf{83.4}$^{\textcolor{red}{+0.9}}$ \\
MISA          & 40.2 & 81.8 & 81.9 \\
MISA + \textbf{SuCI}    & \textbf{41.6}$^{\textcolor{red}{+1.4}}$ & \textbf{83.3}$^{\textcolor{red}{+1.5}}$ & \textbf{83.1}$^{\textcolor{red}{+1.2}}$ \\
Self-MM       & 41.6 & 83.9 & 84.1 \\
Self-MM + \textbf{SuCI} & \textbf{42.0}$^{\textcolor{red}{+0.4}}$ & \textbf{84.3}$^{\textcolor{red}{+0.4}}$ & \textbf{84.6}$^{\textcolor{red}{+0.5}}$ \\
MMIM          & 41.8 & 84.1 & 84.1 \\
MMIM + \textbf{SuCI}    & \textbf{42.4}$^{\textcolor{red}{+0.6}}$ & \textbf{84.9}$^{\textcolor{red}{+0.8}}$ & \textbf{84.8}$^{\textcolor{red}{+0.7}}$ \\
DMD           & 41.0   & 83.3 & 83.2 \\
DMD + \textbf{SuCI}     & \textbf{42.2}$^{\textcolor{red}{+1.2}}$ & \textbf{84.6}$^{\textcolor{red}{+1.3}}$ & \textbf{84.5}$^{\textcolor{red}{+1.3}}$ \\ \bottomrule
\end{tabular}
}
\caption{Comparison results of different methods and SuCI-based models on the MOSI benchmark. Improved results and corresponding gains compared to vanilla methods are marked in \textbf{bold} and \textcolor{red}{red}, respectively.}
\label{mosi}
\end{table}

\subsection{Benchmarks and Model Zoo}
\textbf{Benchmarks.}
Extensive experiments are conducted on three mainstream MLU benchmarks. Concretely, \textbf{MOSI}~\cite{zadeh2016mosi} is a multimodal human sentiment analysis dataset consisting of 2,199 video segments. The standard data partitioning is 1,284 samples for training, 284 samples for validation, and 686 samples for testing. These samples contain a total of 89 distinct subjects from video blogs. 
Each sample is manually annotated with a sentiment score ranging from -3 to 3.
\textbf{MOSEI}~\cite{zadeh2018multimodal} is a large-scale human sentiment and emotion recognition benchmark containing 22,856 video clips from 1,000 different subjects and 250 diverse topics. Among these samples, 16,326, 1,871, and 4,659 samples are used as training, validation and testing sets. 
The sample annotation protocols used for sentiment analysis are consistent with MOSI. \textbf{UR\_FUNNY}~\cite{hasan2019ur} is a multimodal human humor detection dataset that contains 16,514 video clips from 1,741 subjects collected by the TED portal.  There are 10,598, 2,626, and 3,290 samples in the training, validation, and testing sets. Each sample provides the target punchlines and associated contexts from multimodal utterances to support the detection of subject humor/non-humor in the binary labels.
Following widely adopted implementations~\cite{yang2022disentangled,li2023decoupled}, we use the 7-class accuracy ($Acc_7$), binary accuracy ($Acc_2$), and $F1$ score to evaluate the results on MOSI and MOSEI. The standard binary accuracy ($Acc_2$) is utilized for evaluation on UR\_FUNNY.

\begin{table}[t]
\setlength{\tabcolsep}{12pt}
\renewcommand{\arraystretch}{0.8}
\centering
\resizebox{0.8\linewidth}{!}{%
\begin{tabular}{c|ccc}
\toprule
\textbf{Methods}        & $Acc_7 \uparrow$ (\%) & $Acc_2 \uparrow$ (\%) & $ F1 \uparrow$ (\%)   \\ \midrule
EF-LSTM        & 47.4 & 78.2 & 77.9 \\
LF-LSTM         & 48.8 & 80.6   & 80.6 \\ 
Graph-MFN      & 45.0 & 76.9 & 77.0 \\
RAVEN         & 50.0 & 79.1   & 79.5 \\
MCTN         & 49.6 & 79.8 & 80.6 \\
TCSP          & -   & 82.8 & 82.7   \\
PMR           & 52.5 & 83.3 & 82.6 \\
FDMER         & 53.8 & 83.9 & 83.8 \\ \midrule
MulT          & 51.2 & 82.1 & 81.9 \\
MulT + \textbf{SuCI}    & \textbf{52.4}$^{\textcolor{red}{+1.2}}$ & \textbf{83.2}$^{\textcolor{red}{+1.1}}$ & \textbf{82.9}$^{\textcolor{red}{+1.0}}$ \\
MISA         & 51.3 & 82.3 & 82.3 \\
MISA + \textbf{SuCI}    & \textbf{52.6}$^{\textcolor{red}{+1.3}}$ & \textbf{83.5}$^{\textcolor{red}{+1.2}}$ & \textbf{83.2}$^{\textcolor{red}{+0.9}}$ \\
Self-MM       & 52.9 & 83.9 & 83.8 \\
Self-MM + \textbf{SuCI} & \textbf{53.6}$^{\textcolor{red}{+0.7}}$ & \textbf{84.2}$^{\textcolor{red}{+0.3}}$ & \textbf{84.2}$^{\textcolor{red}{+0.4}}$ \\
MMIM          & 53.7 & 84.4 & 84.3 \\
MMIM + \textbf{SuCI}    & \textbf{54.4}$^{\textcolor{red}{+0.7}}$ & \textbf{85.5}$^{\textcolor{red}{+1.1}}$ & \textbf{85.2}$^{\textcolor{red}{+0.9}}$ \\
DMD           & 53.5   & 84.1 & 84.0 \\
DMD + \textbf{SuCI}     & \textbf{54.6}$^{\textcolor{red}{+1.1}}$ & \textbf{85.8}$^{\textcolor{red}{+1.7}}$ & \textbf{85.7}$^{\textcolor{red}{+1.7}}$ \\ \bottomrule
\end{tabular}
}
\caption{Comparison results of different models on the MOSEI benchmark.}
\label{mosei}
\end{table}

\noindent \textbf{Model Zoo.}
Considering the applicability, we choose five representative models with different network structures to verify the proposed plug-in SuCI. A brief overview is as follows. \textbf{MulT}~\cite{tsai2019multimodal} constructs multimodal transformers to learn element dependencies between pairwise modalities. \textbf{MISA}~\cite{Hazarika2020mm} captures modality-invariant and modality-specific diverse representations based on feature disentanglement. \textbf{Self-MM}~\cite{yu2021learning} utilizes a self-supervised paradigm to learn additional emotion semantics from unimodal label generation. \textbf{MMIM}~\cite{han2021improving} proposes a hierarchical mutual information maximization to alleviate the loss of valuable task-related clues. \textbf{DMD}~\cite{li2023decoupled} designs cross-modal knowledge distillations to bridge the semantic gap among modalities.

\subsection{Implementation Details}
All models follow consistent feature extraction procedures for fair comparisons.
The text feature extractor is instantiated by pre-trained Glove word embedding tool~\cite{pennington2014glove} to obtain 300-dimensional linguistic vectors. For MOSI \& MOSEI, we use the library Facet~\cite{iMotions} to extract an ensemble of visual features, including 35 facial action units to reflect facial gesture changes.
Meanwhile, Openface~\cite{baltruvsaitis2016openface} is utilized on UR\_FUNNY to extract 75-dimensional features related to facial behaviors and expressions.
The audio feature extraction is executed utilizing the software COVAREP~\cite{degottex2014covarep} to obtain diverse acoustic attributes, where the dimensions on MOSI \& MOSEI and UR\_FUNNY are 74 and 81, respectively.
The word-aligned data points are employed across all benchmarks.
In the SuCI implementation, the hidden dimensions $d_h$ and $d_n$ are set to 64 and 128, respectively.
The size $d_s$ of each subject confounder is 325, 409, and 456 on MOSI, MOSEI, and UR\_FUNNY, respectively.
We implement the selected methods and SuCI on NVIDIA Tesla A800 GPUs utilizing the PyTorch toolbox, where other training settings are aligned to their original protocols.

\begin{table}[t]
\setlength{\tabcolsep}{12pt}
\renewcommand{\arraystretch}{0.9}
\centering
\resizebox{0.8\linewidth}{!}{%
\begin{tabular}{c|cc|c}
\toprule
\textbf{Methods}        & \textbf{Context} & \textbf{Punchline} & $Acc_2 \uparrow$ (\%)  \\ \midrule
C-MFN         & \checkmark        &           & 58.45 \\
C-MFN        &         & \checkmark          & 64.47 \\
TFN           &         & \checkmark          & 64.71 \\
LMF          &         & \checkmark          & 65.16 \\
C-MFN         & \checkmark        & \checkmark          & 65.23 \\
FDMER         &         & \checkmark          & 70.55 \\ \midrule
MulT         &         & \checkmark          & 66.65 \\
MulT + \textbf{SuCI}    &         & \checkmark          & \textbf{67.88}$^{\textcolor{red}{+1.23}}$ \\
MISA          &         & \checkmark          & 67.34 \\
MISA + \textbf{SuCI}    &         & \checkmark          & \textbf{68.96}$^{\textcolor{red}{+1.62}}$ \\
Self-MM        &         & \checkmark          & 68.77 \\
Self-MM + \textbf{SuCI} &         & \checkmark          & \textbf{69.72}$^{\textcolor{red}{+0.95}}$ \\
MMIM         &         & \checkmark          & 69.53 \\
MMIM + \textbf{SuCI}    &         & \checkmark          & \textbf{70.92}$^{\textcolor{red}{+1.39}}$ \\
DMD           &         & \checkmark          & 68.70  \\
DMD + \textbf{SuCI}     &         & \checkmark          & \textbf{70.84}$^{\textcolor{red}{+2.14}}$ \\ \bottomrule
\end{tabular}
}
\caption{Comparison results of different models on the UR\_FUNNY benchmark.}
\label{funny}
\end{table}

\subsection{Comparison with State-of-the-art Methods}

We compare the SuCI-based models with extensive SOTA methods, including EF-LSTM,  LF-LSTM, C-MFN~\cite{hasan2019ur}, Graph-MFN~\cite{zadeh2018multimodal}, TFN~\cite{zadeh2017tensor}, MFM~\cite{tsai2018learning}, RAVEN~\cite{wang2019words}, MCTN~\cite{pham2019found}, TCSP~\cite{wu2021text}, PMR~\cite{lv2021progressive}, and FDMER~\cite{yang2022disentangled}. 

\noindent \textbf{Results on MOSI Benchmark.}
\textbf{(i)} From Table~\ref{mosi}, SuCI consistently improves the performance of the selected methods on all metrics. Concretely, the overall gains of $Acc_7$,  $Acc_2$, and  $F1$ scores among the five models increased by 4.8\%, 4.8\%, and 4.6\%, respectively. These improvements confirm that our module can break through the performance bottlenecks of most baselines in a model-agnostic manner.
\textbf{(ii)} SuCI can bring more favorable gains for decoupled learning-based efforts.  For instance, MISA and DMD achieve salient relative improvements among different metrics of 1.8\%$\sim$3.5\% and 1.6\%$\sim$2.9\%, respectively. A reasonable explanation is that the decoupling pattern diffuses the spurious semantics caused by subject confounders in multiple feature subspaces. In this case, SuCI's de-confounding ability is more effective.
\textbf{(iii)} Compared to recent PMR and FDMER with complex network stacking and massive parameters~\cite{lv2021progressive,yang2022disentangled}, MMIM achieves the best results by equipping our SuCI.

\begin{table}[t]
\setlength{\tabcolsep}{12pt}
\renewcommand{\arraystretch}{0.9}
\centering
\resizebox{0.8\linewidth}{!}{%
\begin{tabular}{c|ccc}
\toprule
\textbf{Methods}        & $Acc_7 \uparrow$ (\%) & $Acc_2 \uparrow$ (\%) & $ F1 \uparrow$ (\%)   \\ \midrule
MulT           & 46.9 & 77.6 & 77.4 \\
MulT + \textbf{SuCI}    & \textbf{48.5}$^{\textcolor{red}{+1.6}}$ & \textbf{80.2}$^{\textcolor{red}{+2.6}}$ & \textbf{80.3}$^{\textcolor{red}{+2.9}}$ \\
MISA          & 46.6 & 77.2 & 77.1 \\
MISA + \textbf{SuCI}    & \textbf{48.3}$^{\textcolor{red}{+1.7}}$ & \textbf{78.9}$^{\textcolor{red}{+1.7}}$ & \textbf{79.4}$^{\textcolor{red}{+2.3}}$ \\
Self-MM       & 47.7 & 78.5 & 78.3 \\
Self-MM + \textbf{SuCI} & \textbf{48.2}$^{\textcolor{red}{+0.5}}$ & \textbf{79.5}$^{\textcolor{red}{+1.0}}$ & \textbf{79.6}$^{\textcolor{red}{+1.3}}$ \\
MMIM          & 49.3 & 79.6 & 79.3 \\
MMIM + \textbf{SuCI}    & \textbf{51.8}$^{\textcolor{red}{+2.5}}$ & \textbf{82.5}$^{\textcolor{red}{+2.9}}$ & \textbf{82.4}$^{\textcolor{red}{+3.1}}$ \\
DMD            & 48.9   & 80.8 & 80.7 \\
DMD + \textbf{SuCI}     & \textbf{51.6}$^{\textcolor{red}{+2.7}}$ & \textbf{82.4}$^{\textcolor{red}{+1.6}}$ & \textbf{82.4}$^{\textcolor{red}{+1.7}}$ \\ \bottomrule
\end{tabular}
}
\caption{Cross-dataset evaluation of models trained on the MOSI training set and tested on the MOSEI testing set. }
\label{mosi-mosei}
\end{table}

\begin{table}[t]
\setlength{\tabcolsep}{12pt}
\renewcommand{\arraystretch}{0.9}
\centering
\resizebox{0.8\linewidth}{!}{%
\begin{tabular}{c|ccc}
\toprule
\textbf{Methods}        & $Acc_7 \uparrow$ (\%) & $Acc_2 \uparrow$ (\%) & $ F1 \uparrow$ (\%)   \\ \midrule
MulT          & 37.4 & 80.2 & 79.9 \\
MulT + \textbf{SuCI}    & \textbf{38.7}$^{\textcolor{red}{+1.3}}$ & \textbf{81.7}$^{\textcolor{red}{+1.5}}$ & \textbf{81.5}$^{\textcolor{red}{+1.6}}$ \\
MISA          & 37.8 & 80.5 & 80.7 \\
MISA + \textbf{SuCI}    & \textbf{39.0}$^{\textcolor{red}{+1.2}}$ & \textbf{82.2}$^{\textcolor{red}{+1.7}}$ & \textbf{82.3}$^{\textcolor{red}{+1.6}}$ \\
Self-MM       & 38.9 & 82.0 & 82.1 \\
Self-MM + \textbf{SuCI} & \textbf{39.5}$^{\textcolor{red}{+0.6}}$ & \textbf{82.7}$^{\textcolor{red}{+0.7}}$ & \textbf{82.4}$^{\textcolor{red}{+0.3}}$ \\
MMIM          & 39.3 & 82.5 & 82.5 \\
MMIM + \textbf{SuCI}    & \textbf{41.1}$^{\textcolor{red}{+1.8}}$ & \textbf{83.4}$^{\textcolor{red}{+0.9}}$ & \textbf{83.3}$^{\textcolor{red}{+0.8}}$ \\
DMD          & 38.6   & 81.6 & 81.4 \\
DMD + \textbf{SuCI}     & \textbf{40.6}$^{\textcolor{red}{+2.0}}$ & \textbf{83.0}$^{\textcolor{red}{+1.4}}$ & \textbf{82.9}$^{\textcolor{red}{+1.5}}$ \\ \bottomrule
\end{tabular}
}
\caption{Cross-dataset evaluation of models trained on the MOSEI training set and tested on the MOSI testing set. }
\label{mosei-mosi}
\end{table}

\noindent \textbf{Results on MOSEI Benchmark.}
Table~\ref{mosei} provides performance comparison results on MOSEI. 
\textbf{(i)} The SuCI-based models outperform the vanilla methods by large margins
on all metrics. For example, SuCI helps the latest DMD to achieve new SOTA performance with considerable absolute gains of 1.1\%, 1.7\%, and 1.7\% in $Acc_7$,  $Acc_2$, and  $F1$ scores, respectively.
These observations show the broad applicability and usefulness of our module in removing the subject-related prediction bias.
\textbf{(ii)} The improvements provided by SuCI on MOSEI are more significant than those on MOSI. The plausible deduction is that MOSEI contains richer subjects and their highly idiosyncratic multimodal utterances in various scenarios. Thus, SuCI can more accurately remove spurious correlations caused by appropriately extracted subject confounders and offer sufficient gains.

\noindent \textbf{Results on UR\_FUNNY Benchmark.}
From Table~\ref{funny}, we show the detection results using the target punchline for fair comparisons with other works. \textbf{(i)} The SuCI-based models achieve consistent performance increases and accomplish competitive and better results than previous methods. These substantial improvements imply the necessity of performing subject de-confounded training in detecting human humor.
\textbf{(ii)} In particular, MMIM and DMD equipped with SuCI yield absolute gains of 1.39\% and 2.14\%, achieving new SOTAs with the $Acc_2$ of 70.92\% and 70.84\%, respectively.

\subsection{Cross-dataset Evaluation}
Since the used MOSI and MOSEI benchmarks have the same annotation protocols, we establish cross-dataset evaluations for MOSI-training$\rightarrow$MOSEI-testing and MOSEI-training$\rightarrow$MOSI-testing in Tables~\ref{mosi-mosei} and \ref{mosei-mosi}, respectively.
The design intuition is that exploring prediction performance on testing data with different distributions than the training data (\textit{i.e.}, out-of-distribution, OOD) helps verify confounding effects and model generalizability.
The five vanilla methods show severe performance deterioration compared to the results in the Independent Identically Distributed (IID) setting from Tables~\ref{mosi} and \ref{mosei}.
For instance, the testing results on MOSEI and MOSI decreased by the average $Acc_7$ of 4.64\% and 2.42\% across all methods. This is inevitable since spurious correlations between trained models and specific subjects are exacerbated and amplified in uninitiated subjects under the OOD setting.
Fortunately, our SuCI significantly improves the results of all models in cross-dataset evaluations.
These substantial gains confirm that our module favorably mitigates the subject variation problem and enhances the generalizability and robustness of the vanilla models.

\subsection{Ablation Studies}
In Table~\ref{abl}, we perform ablation studies to evaluate the impact of all components in SuCI.
We report results for the $Acc_2$ metrics due to similar trends for the other metrics.

\noindent \textbf{Necessity of Subject Feature Learning.}
\textbf{(i)} We first replace our dynamic fusion mechanism with the average pooling operation along the frame length to obtain refined representations $\bm{p}_m$. The insufficient gains on all benchmarks suggest that assigning adaptive weights based on the distinct frame element contributions in multimodal sequences facilitates better capturing subject-related semantics.
\textbf{(ii)} Subsequently, we separately remove the subject and task discriminators to investigate the role of feature disentanglement. Intuitively, the subject discriminator provides substantial gains for both methods since it supervises the generator to yield discriminative subject features $\bm{s}$ that are better used for confounder construction. 
\textbf{(iii)} The task discriminator is equally critical as it ensures that task-related information is excluded from $\bm{s}$ to distill the pure subject bias.

\begin{table}[t]
\renewcommand{\arraystretch}{0.9}
\centering
\resizebox{0.9\linewidth}{!}{%
\begin{tabular}{ccccccc}
\toprule
\multicolumn{1}{c|}{\multirow{2}{*}{\textbf{Setting}}}     & \multicolumn{2}{c|}{MOSI}                          & \multicolumn{2}{c|}{MOSEI}                         & \multicolumn{2}{c}{UR\_FUNNY}       \\ \cline{2-7} 
\multicolumn{1}{c|}{}                             & \rule{0pt}{10pt}MISA        & \multicolumn{1}{c|}{DMD}           & MISA          & \multicolumn{1}{c|}{DMD}           & MISA           & DMD           \\ \midrule
\multicolumn{1}{c|}{Vanilla  Method}              & 81.8          & \multicolumn{1}{c|}{83.3}          & 82.3          & \multicolumn{1}{c|}{84.1}          & 67.34          & 68.70           \\
\multicolumn{1}{c|}{+ \textbf{The Proposed SuCI}}                       & \textbf{83.3} & \multicolumn{1}{c|}{\textbf{84.6}} & \textbf{83.5} & \multicolumn{1}{c|}{\textbf{85.8}} & \textbf{68.96} & \textbf{70.84} \\ \midrule
\multicolumn{7}{c}{Necessity of Subject  Feature Learning}                                                                                                                                   \\ \midrule
\multicolumn{1}{c|}{w/o Dynamic Fusion Mechanism} & 82.9          & \multicolumn{1}{c|}{84.2}          & 83.2          & \multicolumn{1}{c|}{85.3}          & 68.64          & 70.67          \\
\multicolumn{1}{c|}{w/o Subject Discriminator}    & 82.2          & \multicolumn{1}{c|}{83.8}          & 82.7          & \multicolumn{1}{c|}{85.0}            & 68.57          & 69.69          \\
\multicolumn{1}{c|}{w/o Task Discriminator}       & 82.6          & \multicolumn{1}{c|}{84.1}          & 82.9          & \multicolumn{1}{c|}{85.3}          & 68.62          & 70.44          \\ \midrule
\multicolumn{7}{c}{Importance of  Different Modalities}                                                                                                                                      \\ \midrule
\multicolumn{1}{c|}{w/o Text Modality}            & 82.4          & \multicolumn{1}{c|}{83.6}          & 82.7          & \multicolumn{1}{c|}{84.6}          & 68.25          & 69.38          \\
\multicolumn{1}{c|}{w/o Visual Modality}          & 82.9            & \multicolumn{1}{c|}{84.3}          & 83.1          & \multicolumn{1}{c|}{85.4}          & 68.62          & 69.94          \\
\multicolumn{1}{c|}{w/o Audio Modality}           & 82.7          & \multicolumn{1}{c|}{84.1}          & 83.0            & \multicolumn{1}{c|}{85.1}          & 68.77          & 70.35          \\ \midrule
\multicolumn{7}{c}{Rationality of Confounder Dictionary}                                                                                                                                       \\ \midrule
\multicolumn{1}{c|}{w/ Random $\bm{Z}$}                  & 79.3          & \multicolumn{1}{c|}{81.0}            & 78.2          & \multicolumn{1}{c|}{79.9}          & 62.46          & 84.05          \\
\multicolumn{1}{c|}{w/ Clustered $\bm{Z}$}               & 82.8          & \multicolumn{1}{c|}{83.7}          & 82.9          & \multicolumn{1}{c|}{85.0}          & 67.98          & 69.63          \\ \midrule
\multicolumn{7}{c}{Effectiveness of  Adaptive Aggregation}                                                                                                                                   \\ \midrule
\multicolumn{1}{c|}{w/o $\psi_{i}$}                      & 82.7          & \multicolumn{1}{c|}{84.2}          & 83.1          & \multicolumn{1}{c|}{85.4}          & 68.45          & 70.32          \\
\multicolumn{1}{c|}{w/o $p(\bm{z}_i)$ }                      & 83.0            & \multicolumn{1}{c|}{84.5}          & 83.4          & \multicolumn{1}{c|}{85.6}          & 68.73          & 70.66          \\ \bottomrule
\end{tabular}
}
\caption{Ablation studies on the three benchmarks. ``w/'' and ``w/o'' mean the with and without.}
\label{abl}
\end{table}

\noindent \textbf{Importance of  Different Modalities.}
\textbf{(i)} When the text, visual, and audio modalities from the subjects are removed separately in de-confounded training, the improvements in SuCI for the baselines show significant deterioration.
These decreased results confirm that subject-specific spurious characteristics are transmitted in multimodal utterances and that considering the complete modalities is necessary.
\textbf{(ii)} The reason for the pronounced effect of the text modality may be the adverse statistical shortcuts caused by the highly uneven distribution of textual words across the samples of distinct subjects, which is a confounder inducer.

\begin{figure}[t]
  \centering
  \includegraphics[width=0.7\linewidth]{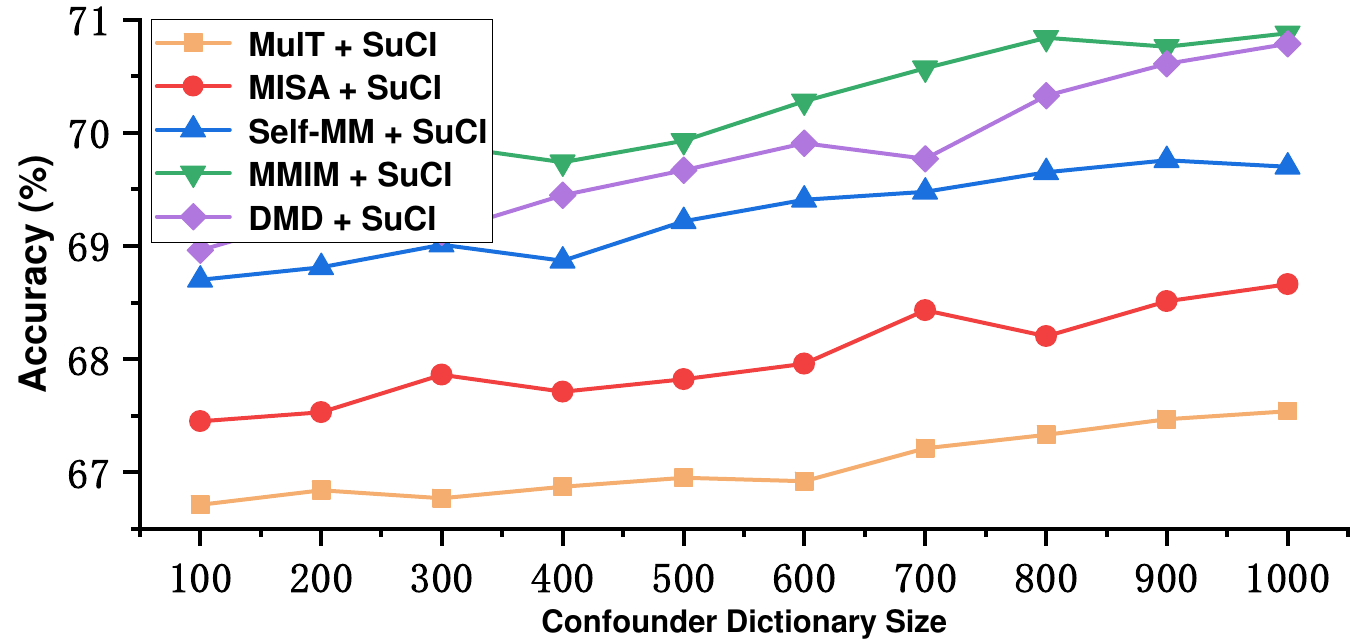}
  \caption{Ablation study results for the number of subject confounders on the UR\_FUNNY benchmark.
  }
\label{size}
\end{figure}

\begin{figure}[t]
  \centering
  \includegraphics[width=0.8\linewidth]{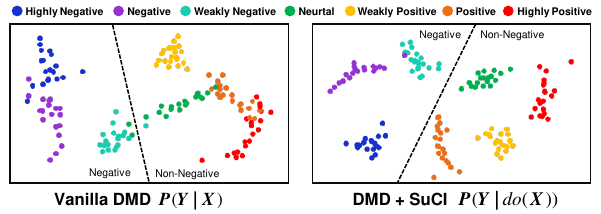}
  \caption{Quantitative results (\textit{i.e.}, binary or seven classifications) of vanilla and SuCI-based DMD on the MOSEI.
  }
\label{vis-mosei}
\end{figure}

\noindent \textbf{Rationality of Confounder Dictionary.}
We provide two candidates of the same size as the default confounder dictionary $\bm{Z}$ to evaluate the rationality of our confounder construction. These two alternative dictionaries are obtained by random initialization and unsupervised K-Means++ clustering~\cite{bahmani2012scalable}.
\textbf{(i)} The random $\bm{Z}$ would significantly impair the performance and even underperform the vanilla methods, justifying the proposed subject prototypes as confounders.
\textbf{(ii)} The performance gains of the clustered $\bm{Z}$ are sub-optimal due to the lack of subject-specific information supervision, leading to the indistinguishability and coupling of confounding effects across different subjects.

\noindent \textbf{Effectiveness of Adaptive Aggregation.}
Here, the adaptive aggregation strategy is explored by eliminating the attention weights $\psi_{i}$ and the prior probabilities $p(\bm{z}_i)$ in $\mathbb{E}[h(\bm{z})]$, respectively. The consistent performance drops on all benchmarks imply that characterizing the importance and proportion of each subject confounder is indispensable for achieving effective causal intervention based on subject debiasing.

\noindent \textbf{Impact of Subject Confounder Number.}
Finally, we test the impact of different numbers of subject confounders on SuCI performance in Figure~\ref{size}.
The SuCI-based models all exhibit overall rising gain trends as the training subjects increase.
These findings suggest that sufficient stratified confounders facilitate better backdoor adjustment implementation and accurate average causal effect estimation.

\subsection{Qualitative Evaluation}
To show the difference between the model approximate $P(\bm{Y}|\bm{X})$ and $P(\bm{Y}|do(\bm{X}))$, we randomly select several samples in each sentiment category on MOSEI for visualization.
As Figure~\ref{vis-mosei} shows, the sample distributions from the different categories of the vanilla baseline prediction usually appear confounded and fragmented. In comparison, the causality-based SuCI improves the predicted results so that the samples in the same category are more compact and different categories are well separated.
The observation suggests that our causal intervention promotes discriminative predictions by mitigating the subject prediction bias.

\section{Conclusion}
This paper is the first to reveal the long-neglected subject variation problem in MLU tasks and identify subjects as essentially harmful confounders from a causal perspective. 
Thus, we present a subject causal intervention module (SuCI) to remove the prediction bias caused by the subject-specific spurious correlations. Extensive experiments show the broad applicability of SuCI in diverse methods.

\section{Acknowledgments}
This work is supported in part by the National Key R\&D Program of China (2021ZD0113503).

\bibliography{aaai25}

\end{document}